\documentclass{jpsj3}
\usepackage{bm}
\usepackage{bm}
\usepackage{amssymb}

\title{Inverse Renormalization Group Transformation in Bayesian Image Segmentations}

\author{
\name{Kazuyuki Tanaka
\thanks{E-mail: kazu@smapip.is.tohoku.ac.jp}}$^{1}$, 
\name{Shun Kataoka}$^{1}$, 
\name{Muneki Yasuda}$^{2}$, 
\name{Masayuki Ohzeki}$^{3}$ \\
}
\inst{
\address{$^{1}$Graduate School of Information Sciences,
Tohoku University, 6-3-09 Aramaki-aza-aoba,
Aoba-ku, Sendai 980-8579, Japan}\\
\address{$^{2}$Graduate School of Science and Engineering, 
Yamagata University, 
4-3-16 Jyounan, Yonezawa 992-8510, Japan}\\
\address{$^{3}$Graduate School of Informatics,
Kyoto University, 36-1 Yoshida-Honmachi, Sakyo-ku, Kyoto 606-8501 Japan}\\
} %\\

\abst{
A new Bayesian image segmentation algorithm is proposed 
by combining a loopy belief propagation 
with an inverse real space renormalization group transformation
to reduce the computational time.
In results of our experiment, we observe that the proposed method 
can reduce the computational time to less than one-tenth 
of that taken by conventional Bayesian approaches.
}

\kword{statistical-mechanical informatics,
Markov random fields, renormalization group}

\begin{document}

\maketitle

Bayesian segmentation modeling based 
on Markov random fields (MRF's) 
is one of the interesting research topics{\cite{KatoZerubia2011}}.
Image segmentations are required to classify pixels 
in an observed image 
into  several regions, and such segmentations 
are regarded as a kind of clustering of pixels.
Markov random fields are regarded as classical spin systems 
in statistical mechanics{\cite{KTanaka2002}}. 
Loopy belief propagations (LBP's) 
have been applied to construct 
certain practical algorithms 
for application in Bayesian image 
segmentations{\cite{HasegawaOkadaMiyoshi2011,
TanakaKataokaYasudaWaizumiHsu2014}}.
In the present short note, we propose 
a new Bayesian image segmentation algorithm 
based on a RSRG transformation
to reduce the computational time.

We consider an image 
as defined on a set of pixels 
arranged on a square grid graph $({\cal{V}},{\cal{E}})$. 
Here ${\cal{V}}{\equiv}\{i|i=1,2,{\cdots},|{\cal{V}}|\}$ denotes 
the set of all the pixels 
and ${\cal{E}}$ is the set of 
all the nearest-neighbour pairs 
of pixels $\{i,j\}$. 
The total numbers of elements 
in the sets ${\cal{V}}$ and ${\cal{E}}$ 
are denoted by $|{\cal{V}}|$ and $|{\cal{E}}|$, respectively.
The square grid graph has the periodic boundary conditions 
along the $x$- and $y$-directions.
The label at each pixel $i$ is regarded as a state variable, 
and it is denoted by $a_{i}$.
Each pixel $i$ takes on the values of all the possible integers 
in the set ${\cal{Q}}{\equiv}\{0,1,2,{\cdots},q-1\}$ 
as its region label. 
The state vector of labels is represented 
by ${\bm{a}}=(a_{i}|i{\in}{\cal{V}})
=(a_{1},a_{2},{\cdots},a_{|{\cal{V}}|})^{\rm{T}}$. 
The prior probability 
of a labeling configuration ${\bm{a}}$ 
is assumed to be specified by a constant $u$ as
\begin{eqnarray}
P({\bm{a}})
\propto
{\prod_{\{i,j\}{\in}{\cal{E}}}}
{\exp}{\Big (}{\frac{1}{2}}{\alpha}
{\delta}_{a_{i},a_{j}}{\Big )},
\label{Prior}
\end{eqnarray}
up to the normalization constant. 

\begin{figure}
\begin{center}
\includegraphics[height=7.0cm, bb = 0 50 566 488]{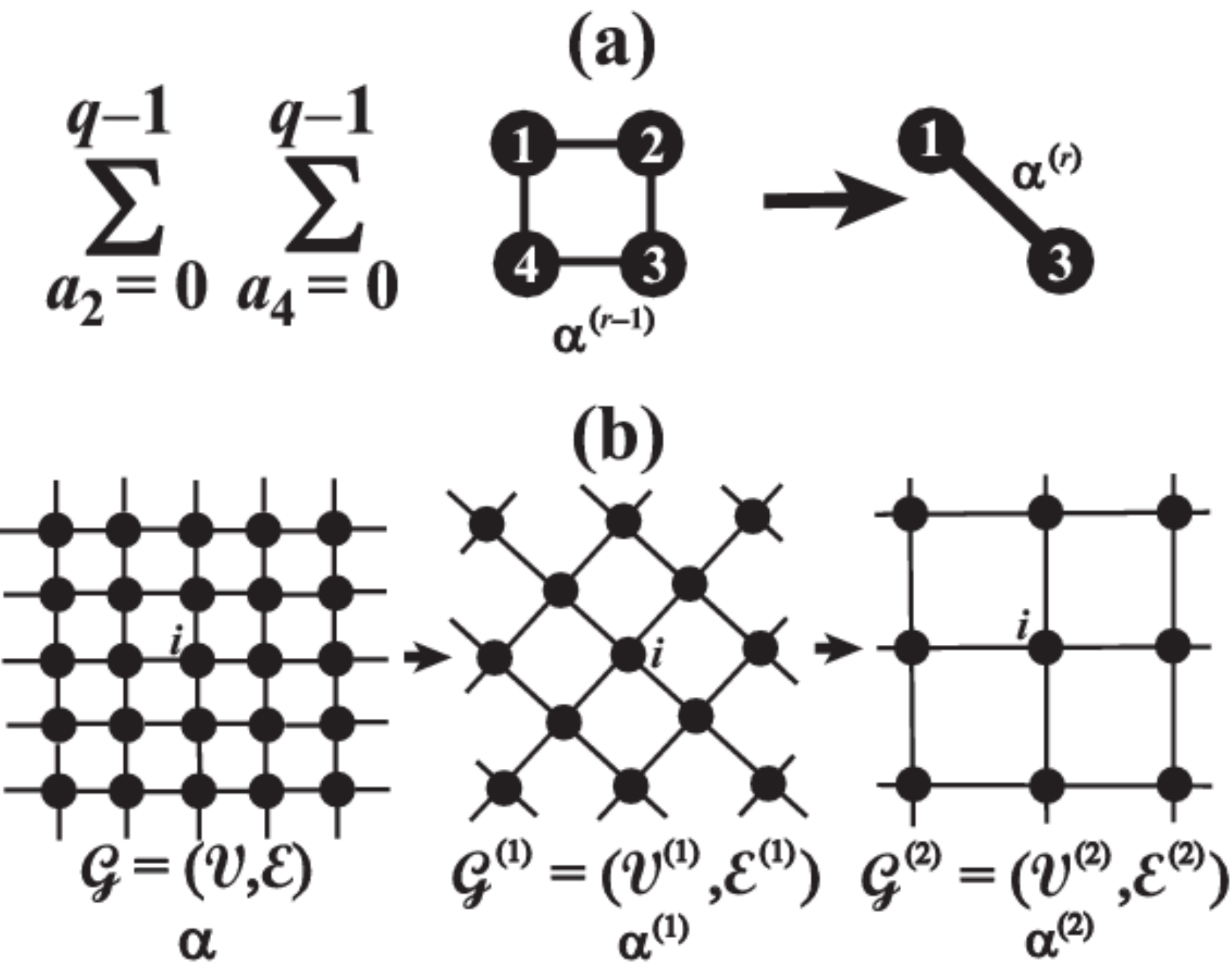}
\end{center}
\caption{
Block spin transformation.
(a)~Graphical representation of Eq.(\ref{BlockSpinTransformation}).
(b)~${\cal{G}}^{(r)}=({\cal{V}}^{(r)},{\cal{E}}^{(r)})$ ($r=1,2,{\cdots}$) 
in the coarse-graining process.
}
\label{Figure01}
\end{figure}

For the prior probability distribution $P({\bm{a}}|u)$, 
we introduce the following RSRG transformation:
\begin{eqnarray}
{\exp}{\Big (}{\frac{1}{2}}{\alpha}^{(r)}
{\delta}_{a_{1},a_{3}}{\Big )}
& \varpropto &
{\sum_{a_{2}{\in}{\cal{Q}}}}
{\sum_{a_{4}{\in}{\cal{Q}}}}
{\exp}{\Big (}{\frac{1}{2}}{\alpha}^{(r-1)}
{\big (}{\delta}_{a_{1},a_{2}}
+{\delta}_{a_{2},a_{3}}
+{\delta}_{a_{1},a_{4}}
+{\delta}_{a_{4},a_{3}}{\big )}{\Big )}
\nonumber\\
& &{\hspace{3.0cm}}
(r=1,2,{\cdots},R),
\label{BlockSpinTransformation}
\end{eqnarray}
where ${\alpha}^{(0)}{\equiv}{\alpha}$. 
The square grid graph 
${\cal{G}}^{(r)}=({\cal{V}}^{(r)},{\cal{E}}^{(r)})$ 
is defined as shown in Fig.{\ref{Figure01}}(b).
Here ${\cal{V}}^{(r)}$ is the set of all the pixels 
and ${\cal{E}}^{(r)}$ represents the set of 
all the nearest-neighbour pairs 
of pixels $\{i,j\}$ in the graph ${\cal{G}}^{(r)}$.
Here, we remark that 
${\cal{V}}^{(r)}$ is the subset of ${\cal{V}}$. 
The prior probability distribution 
$P^{(r)}({\bm{a}}^{(r)})$ 
for ${\bm{a}}^{(r)}{\equiv}(a_{i}|i{\in}{\cal{V}}^{(r)})$ 
on ${\cal{G}}^{(r)}=({\cal{V}}^{(r)},{\cal{E}}^{(r)})$
after the $r$-iteration of the RSRG transformation 
is expressed as 
\begin{eqnarray}
P^{(r)}({\bm{a}}^{(r)})
\propto
{\prod_{\{i,j\}{\in}{\cal{E}}^{(r)}}}
{\exp}{\Big (}{\frac{1}{2}}{\alpha}^{(r)}
{\delta}_{a_{i},a_{j}}{\Big )},
\label{RGPrior}
\end{eqnarray}
up to the normalization constant.
The transformation in Eq.(\ref{BlockSpinTransformation}) 
can be reduced to 
\begin{eqnarray}
{\alpha}^{(r)}=4
{\ln}{\Big (}
{\frac{q-1+e^{{\alpha}^{(r-1)}}}
{q-2+2e^{{\frac{1}{2}}{\alpha}^{(r-1)}}}}
{\Big )}.
\label{RGTransformationA}
\end{eqnarray}

The intensities of 
the red, green, and blue channels at pixel $i$ 
in the observed image 
are regarded as state variables 
denoted by 
$d_{i}^{\rm{Red}}$, $d_{i}^{\rm{Green}}$ 
and $d_{i}^{\rm{Blue}}$, respectively.
The random fields of red, green and blue intensities 
in the observed color image are then represented
by the $3|{\cal{V}}|$-dimensional vector 
${\bm{d}}{\equiv}({\bm{d}}_{1},{\bm{d}}_{2},
{\cdots},{\bm{d}}_{|{\cal{V}}|})^{\rm{T}}$, 
where ${\bm{d}}_{i}
\equiv (d_{i}^{\rm{Red}},d_{i}^{\rm{Green}},
d_{i}^{\rm{Blue}})^{\rm{T}}$.
The state variables $d_{i}^{\rm{Red}}$, $d_{i}^{\rm{Green}}$ 
and $d_{i}^{\rm{Blue}}$ at each pixel $i$ can take 
any real numbers 
in the interval $(-{\infty},+{\infty})$.
It is to be noted that, in the above generative process of natural color images, 
we assign each pixel $i$ to a labeling state $a_{i}$.
For example, if the pixel $i$ is 
in the labeling state $a_{i}={\xi}$ (${\xi}{\in}{\cal{Q}}$),
then its color intensity vector ${\bm{d}}_{i}$ is 
assumed to be generated 
from the following Gaussian distribution: 
\begin{eqnarray}
g({\bm{d}}_{i}|{\xi})
\equiv 
{\sqrt{{\frac{1}{{\rm{det}}(2{\pi}{\bm{C}}({\xi}))}}}}
{\exp}{\Big (}
-{\frac{1}{2}}
({\bm{d}}_{i}-{\bm{m}}({\xi}))^{\rm{T}}
{\bm{C}}^{-1}({\xi})
({\bm{d}}_{i}-{\bm{m}}({\xi}))
{\Big )}.
\label{2DGaussian}
\end{eqnarray}
In other words, the labeling state $a_{i}$ 
specifies the distribution within the set 
$\{g({\bm{d}}_{i}|{\xi})|{\xi}{\in}{\cal{Q}}\}$ 
that generates a color intensity vector ${\bm{d}}_{i}$. 
As mentioned in the previous section, 
we introduce 
the labeling state variable $a_{i}$ at each pixel $i$ 
as a Potts spin variable in statistical mechanics.

After setting $R$ as a positive integer, 
we construct a $3|{\cal{V}}^{(R)}|$-dimensional 
vector ${\bm{d}}^{(R)}=({\bm{d}}_{i}|i{\in}{\cal{V}}^{(R)})$ 
from our observed color image ${\bm{d}}$.
We remark that $d_{i}$ 
in the coarse-grained image ${\bm{d}}^{(R)}$ 
is always the same as $d_{i}$ in the observed color image ${\bm{d}}$ 
in our scheme.
By using the Bayes formula, 
we introduce the posterior probability distribution 
$P{\big (}{\bm{a}}^{(R)}{\big |}{\bm{d}}^{(R)}{\big )}$ 
for ${\bm{a}}^{(R)}=(a_{i}|i{\in}{\cal{V}}^{(R)})$ 
as follows:
\begin{eqnarray}
P^{(R)}{\big (}{\bm{a}}^{(R)}{\big |}{\bm{d}}^{(R)}{\big )}
\propto
{\Big (}{\prod_{i{\in}{\cal{V}}^{(R)}}}
g({\bm{d}}_{i}|a_{i}){\Big )}
{\Big (}{\prod_{\{i,j\}{\in}{\cal{E}}^{(R)}}}
{\exp}{\big (}{\frac{1}{2}}{\alpha}^{(R)}
{\delta}_{a_{i},a_{j}}{\big )}{\Big )},
\label{Posterior}
\end{eqnarray}
up to the normalization constant. 
The estimates ${\widehat{\alpha}}^{(R)}$ and 
$\{{\widehat{\bm{m}}}({\xi}),
{\widehat{\bm{C}}}({\xi})|{\xi}{\in}{\cal{Q}}\}$
of ${\alpha}^{(R)}$ and 
$\{{\bm{m}}({\xi}),{\bm{C}}({\xi})|{\xi}{\in}{\cal{Q}}\}$. 
are approximately computed 
by means of the LBP algorithm 
for the conditional maximum entropy 
framework{\cite{TanakaKataokaYasudaWaizumiHsu2014}}.
After obtaining these estimates, 
we calculate ${\widehat{\alpha}}^{(0)}$ 
from ${\widehat{\alpha}}^{(R)}$ 
by using the following inverse RSRG transformations 
of Eq.(\ref{RGTransformationA}):
\begin{eqnarray}
{\widehat{\alpha}}^{(r-1)}
= 2{\ln}{\Big (}
e^{{\frac{1}{4}}{\alpha}^{(r)}}
+
{\sqrt{(e^{{\frac{1}{4}}{\widehat{\alpha}}^{(r)}}+Q-1)
(e^{{\frac{1}{4}}{\widehat{\alpha}}^{(r)}}-1)}}
{\Big )}
~(r=R,{\cdots},2,1),
\label{InverseRGTransformationA}
\end{eqnarray}
where ${\widehat{\alpha}}={\widehat{\alpha}}^{(0)}$. 

Given the estimates 
${\widehat{\alpha}}$ and 
$\{{\widehat{\bm{m}}}({\xi}),
{\widehat{\bm{C}}}({\xi})|{\xi}{\in}{\cal{Q}}\}$,
the estimate of labeling 
${\widehat{\bm{a}}}({\bm{d}})
=({\widehat{a}}_{1}({\bm{d}}),
{\widehat{a}}_{2}({\bm{d}}),
{\cdots},
{\widehat{a}}_{|{\cal{V}}|}({\bm{d}}))^{\rm{T}}$ 
is determined by
\begin{eqnarray}
{\widehat{a}}_{i}({\bm{d}})
\equiv
{\arg}
{\max_{{\zeta}{\in}{\cal{Q}}}}
{\displaystyle{{\sum_{{\bm{a}}}}}}
{\delta}_{a_{i},{\zeta}}
P^{(0)}{\big (}{\bm{a}}{\big |}{\bm{d}}{\big )}
{\hspace{2.0mm}}(i{\in}{\cal{V}}).
\label{MPM}
\end{eqnarray}
This procedure to determine the estimate ${\widehat{a}}({\bm{d}})$ 
is approximately computed 
by using the LBP algorithm to 
$P^{(0)}{\big (}{\bm{a}}{\big |}{\bm{d}}{\big )}$. 

\begin{figure}
\begin{center}
\begin{tabular}{ c c c c }
{\bf{(a)}}
&{\bf{(b)}}
&{\bf{(c)}}
&{\bf{(d)}}
\\
\includegraphics[height=5.0cm, bb = 0 0 192 288]{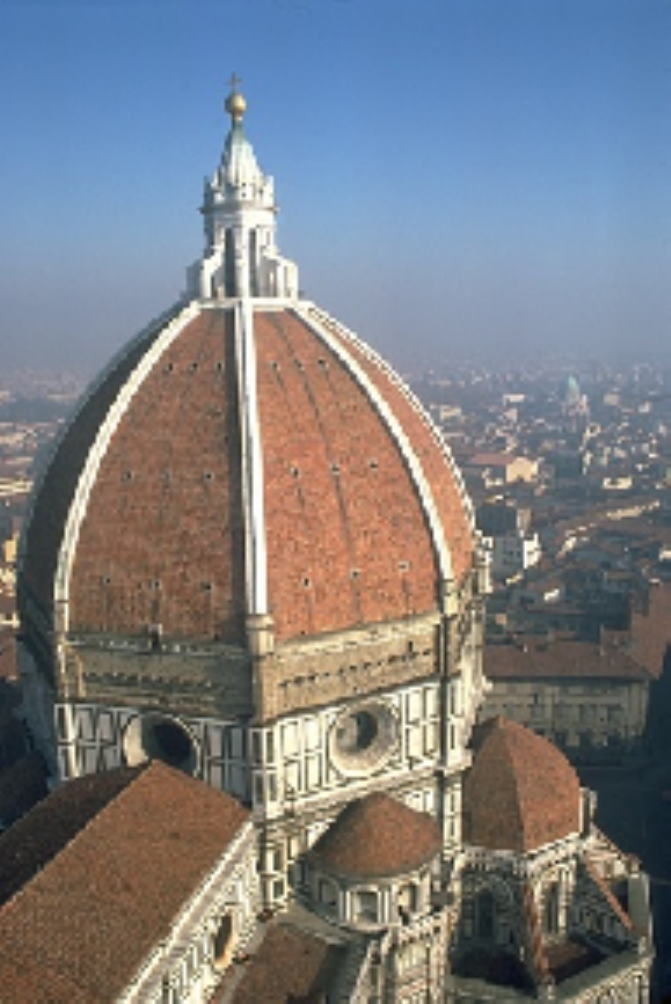}
&\includegraphics[height=5.0cm, bb = 0 0 192 288]{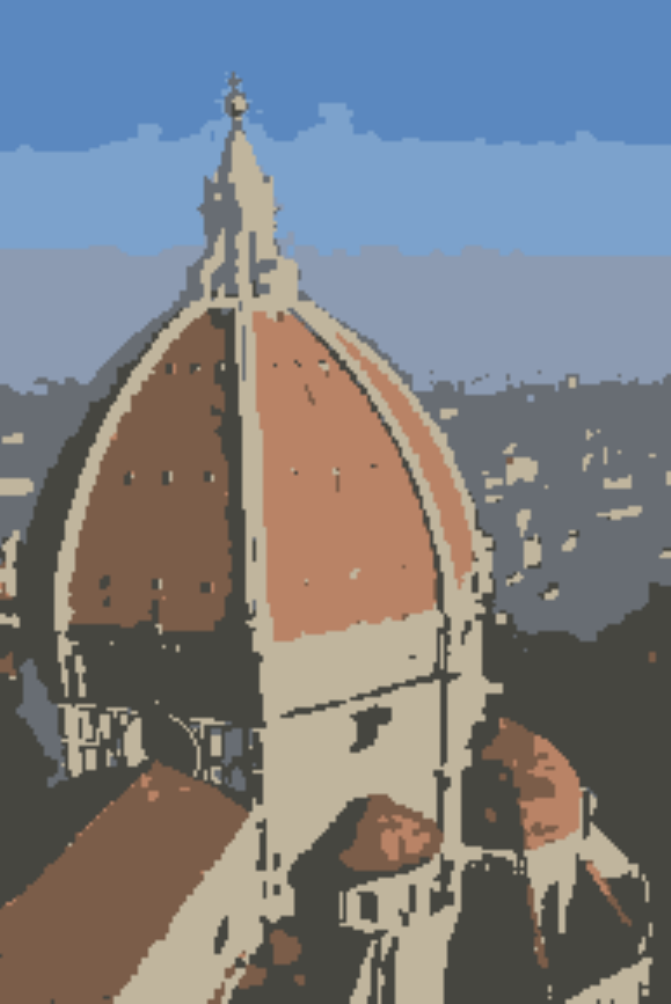}
&\includegraphics[height=5.0cm, bb = 0 0 192 288]{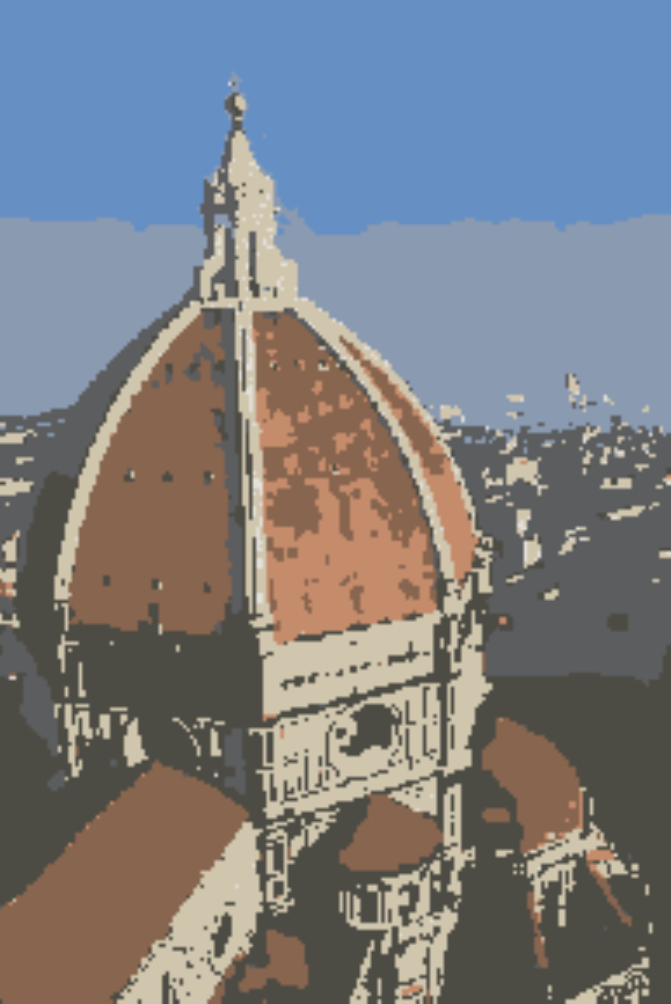}
&\includegraphics[height=5.0cm, bb = 0 0 192 288]{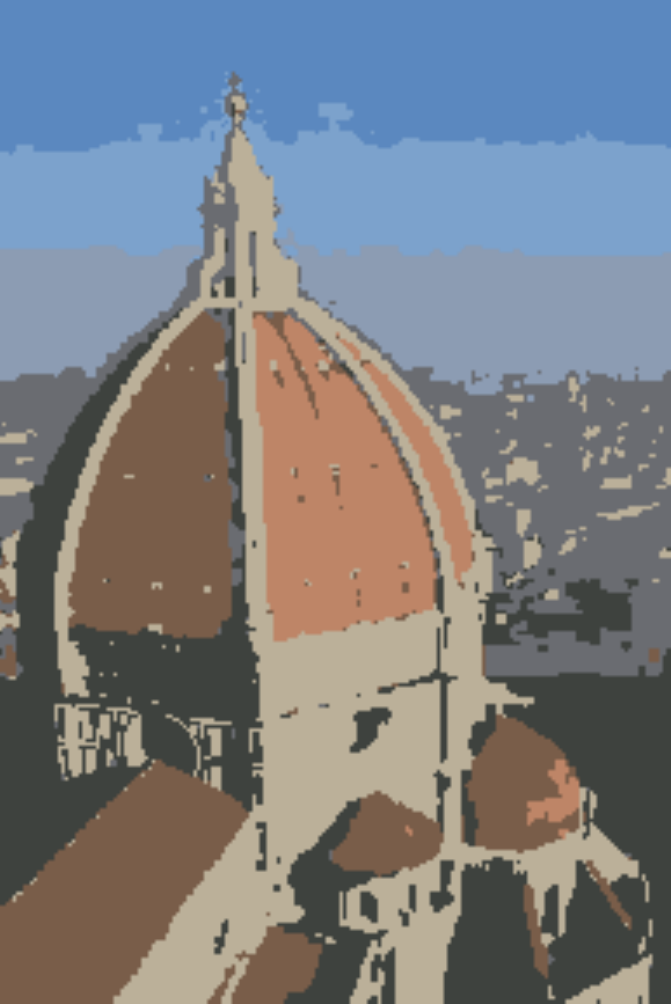}
\end{tabular}
\end{center}
\caption{Image segmentations in the case of $q=8$.
(a) Test image ${\bm{d}}$ from the database 
of Ref.{\cite{ArbelaezMaireFowlkesMalik2011}}. 
(b)-(c) Labeling configurations by our proposed algorithm based 
on our inverse RSRG transformation for (b) $R=8$ and (c) $R=10$. 
(c) Labeling configuration 
by Ref.{\cite{TanakaKataokaYasudaWaizumiHsu2014}} 
without using the inverse RSRG transformation. 
The results in (b)-(d) are shown with the color 
${\widehat{\bm{m}}}{\big (}{\widehat{a}}_{i}({\bm{d}}),{\bm{d}}{\big )}$ 
at each pixel $i$ for the test image ${\bm{d}}$.
}
\label{Figure02}
\end{figure}

We show numerical experiments by our proposed approach in Fig.{\ref{Figure02}}.
In our numerical experiments, the test image ${\bm{d}}$ 
in Fig.{\ref{Figure02}}(a) 
is acquired from the Berkeley Segmentation 
Data Set 500 (BSDS500){\cite{ArbelaezMaireFowlkesMalik2011}}.
The size of the test image ${\bm{d}}$ is $321{\times}481$ 
and the sizes of ${\bm{d}}^{(R)}$ are reduced to $20{\times}30$ for $R=8$ 
and $10{\times}20$ for $R=10$.
The labeling configurations ${\bm{\widehat{a}}}$ 
obtained by means of 
our proposed algorithm based 
on our inverse RSRG transformation for $R=8$ and $R=10$ 
are shown in Figs.{\ref{Figure02}}(b) and (c), respectively.
In our proposed algorithm, 
${\widehat{a}}^{(R)}$ and 
$\{{\widehat{\bm{m}}}({\xi}),
{\widehat{\bm{C}}}({\xi})|{\xi}{\in}{\cal{Q}}\}$ 
for ${\bm{d}}^{(R)}$ are estimated 
by using the conditional maximum entropy framework 
with the LBP 
such that 
they are obtained by subjecting ${\bm{d}}^{(R)}$ 
to the algorithm presented 
in {\S}3 of Ref.{\cite{TanakaKataokaYasudaWaizumiHsu2014}}. 
The estimates ${\widehat{\alpha}}^{(R)}$ for $R=8$ and $=10$ 
are $2.5288$ and $2.5039$, respectively. 
The estimates ${\widehat{\alpha}}={\widehat{\alpha}}^{(0)}$ 
obtained by means of the inverse RSRG group 
transformations given by Eq.(\ref{InverseRGTransformationA}) 
for $R=8$ and $R=10$ 
are $3.6765$ and $3.6797$, respectively.
The labeling configuration in Fig.{\ref{Figure02}}(d) 
is computed directly by applying ${\bm{d}}$ 
to the algorithm in {\S}3 
of Ref.{\cite{TanakaKataokaYasudaWaizumiHsu2014}}. 
The estimate ${\widehat{\alpha}}$ 
of ${\alpha}$ in Fig.{\ref{Figure02}}(d) is $3.0301$.
The computational times corresponding to 
Figs.{\ref{Figure02}}(b)-(d) are 
$128$(Sec), $78$(Sec) and $1331$(Sec), respectively.
Our numerical experiments were performed 
by using a personal computer with an Intel(R) Core(TM) i7-4600U CPU 
with a memory of 8GB.
From the results of our experiment, we observe 
that the computational time can be reduced to less than one-tenth 
of that taken by conventional methods by application 
of our proposed algorithm based on the inverse RSRG transformation.
We obtained results similar to the above mentioned case 
for other test images in the database set of BSDS500. 

In the present short note, we have presented a novel algorithm 
that involve the combination of the inverse RSRG transformation with 
the Bayesian image segmentation method 
proposed in Ref.{\cite{TanakaKataokaYasudaWaizumiHsu2014}}.
Our proposed method can reduce the computational time 
of the hyperparameter estimations significantly.
We expect that our approach can also be applied 
to Bayesian image segmentations for three-dimensional computer vision, 
which remains one of significant problems.

\section*{Acknowledgements}
The authors are grateful to Prof. Federico Ricci-Tersenghi
of the Department of Physics, University of Roma La Sapienza 
for valuable comments.
This work was partly supported 
by the JST-CREST 
and the Grants-In-Aid (No.25280089) 
for Scientific Research from the Ministry of Education, 
Culture, Sports, Science and Technology of Japan.

% \section*{References}


\begin{thebibliography}{9}
\bibitem{KatoZerubia2011}
         Z. Kato and J. Zerubia: 
         Foundations and Trends in Signal Processing, 
         {\bf{5}} (2012) 1.
\bibitem{KTanaka2002} 
         K. Tanaka: 
         J. Phys. A {\bf{35}} (2002) R81.
\bibitem{HasegawaOkadaMiyoshi2011} 
         R. Hasegawa, M. Okada and S. Miyoshi:
         J. Phys. Soc. Jpn {\bf{80}} (2011) 093802.
\bibitem{TanakaKataokaYasudaWaizumiHsu2014}
         K. Tanaka, S. Kataoka, M. Yasuda, Y. Waizumi and C.-T. Hsu:
         J. Phys. Soc. Jpn {\bf{83}} (2014) 124002.
\bibitem{ArbelaezMaireFowlkesMalik2011} 
         P. Arbelaez, M. Maire, C. Fowlkes and J. Malik:
         IEEE Trans. Pattern Anal. Mach. Intell. 
         {\bf{33}} (2011) 898.
\end{thebibliography}
\end{document}